\documentclass[journal]{IEEEtran}
\ifCLASSOPTIONcompsoc
    \usepackage[nocompress]{cite}
\else
    \usepackage{cite}
\fi
\usepackage{amsmath}
\usepackage{amssymb}
\usepackage{comment}
\usepackage{graphicx}
\usepackage{multirow}
\usepackage{ragged2e}
\usepackage[caption=false]{subfig}
\usepackage{tikz}
\usepackage{xcolor}
\usepackage[breaklinks,colorlinks,pagebackref]{hyperref}

\begin{document}

\title{Effects of Degradations on Deep Neural Network Architectures}
\author{Prasun Roy$^*$,~Subhankar Ghosh$^*$,~Saumik~Bhattacharya$^*$,~and~Umapada Pal, \emph{Senior member, IEEE} \\
Indian Statistical Institute Kolkata, India \\
\url{https://github.com/prasunroy/cnn-on-degraded-images}
\thanks{$^*$These authors contributed equally to this work.}
}

\maketitle

\begin{abstract}
Deep convolutional neural networks (CNN) have massively influenced recent advances in large-scale image classification. More recently, a dynamic routing algorithm with capsules (groups of neurons) has shown state-of-the-art recognition performance. However, the behavior of such networks in the presence of a degrading signal (noise) is mostly unexplored. An analytical study on different network architectures toward noise robustness is essential for selecting the appropriate model in a specific application scenario. This paper presents an extensive performance analysis of six deep architectures for image classification on six most common image degradation models. In this study, we have compared VGG-16, VGG-19, ResNet-50, Inception-v3, MobileNet and CapsuleNet architectures on Gaussian white, Gaussian color, salt-and-pepper, Gaussian blur, motion blur and JPEG compression noise models.
\end{abstract}


\IEEEpeerreviewmaketitle

\section{Introduction}\label{sec:introduction}
\IEEEPARstart{V}{isual} quality is an essential evaluation metric in machine vision problems. Though there are several no-reference image quality measures available in literature \cite{mittal2012no}, the visual quality of an image is highly subjective and relies on human judgment. However, computer vision algorithms work differently from the human vision system \cite{ullman2016atoms}, and the context of image quality in computer vision does not always align with human perception. CNNs extract distinctive image features using several convolution filters before classification. Thus, it is difficult to intuitively predict the inference behavior on a degraded image. The classification accuracy mainly depends on the network architecture and degradation model. Generally, we train and validate a network on high-quality images with minimum noise. However, in most practical scenarios, different noise models can introduce degradation of the input image that heavily affects the intended performance of the CNNs. Some common causes of such degradations include poor image sensors, inadequate lighting, improper focus, stabilization problems, wrong exposure and lossy compressions. To minimize such distortions, some researchers \cite{dodge2016understanding} have suggested introducing noisy samples in the training data. Although this technique improves the robustness of the networks against image degradation, it is very challenging to determine every probable noise model that may corrupt the input image.

Convolutional neural networks constitute the core building blocks of most state-of-the-art networks for various machine vision tasks, including classification, object detection, and generation. CNN architectures such as VGG \cite{simonyan2014very}, ResNet \cite{he2016deep}, Inception \cite{szegedy2016rethinking}, and MobileNet \cite{howard2017mobilenets} have achieved exceptional recognition rates in large-scale image classification (ILSVRC). Interestingly, researchers \cite{moosavi2016deepfool} have shown that degrading an image with a small amount of carefully estimated noise can cause severe misclassification by a trained network, even when these distortions are not perceivable to human vision. Although the probability of occurrence of such adversarial perturbations is generally low, it is crucial to study their effects on various deep architectures for building more robust networks.

\begin{figure}[t]
    \includegraphics[width=\linewidth]{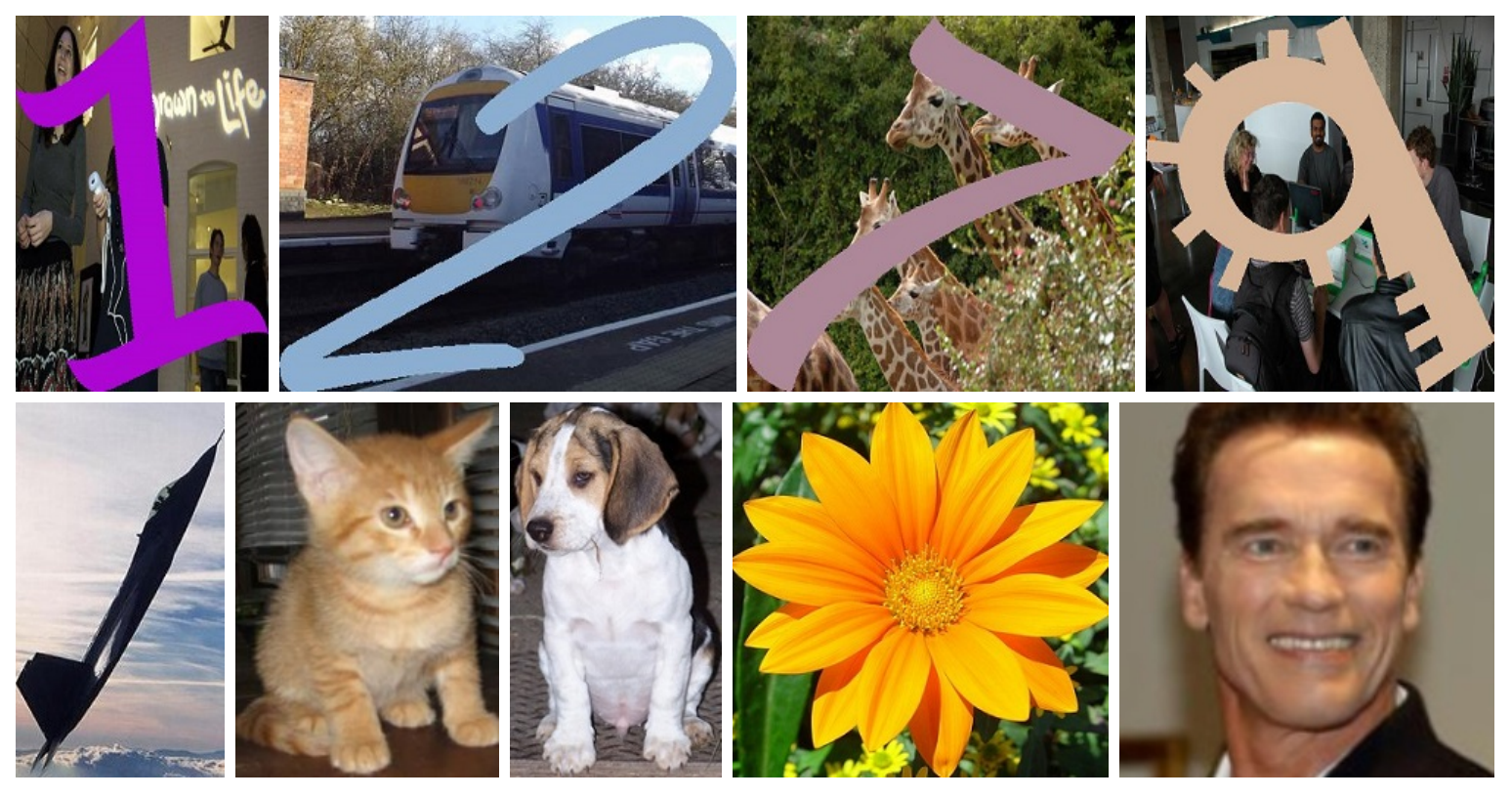}
    \caption{Examples of image samples from the datasets proposed in this work. \textbf{Top row:} Synthetic digits dataset. \textbf{Bottom row:} Natural images dataset.}
    \label{fig:dataset_samples}
\end{figure}

In this paper, we have extensively investigated the classification performance of several deep networks on different image degradation models. The main contributions of this work are as follows.

\begin{itemize}
    \item We have studied the robustness of six state-of-the-art deep architectures on image recognition against six different noise models.
    \item We introduce a modified fusion architecture \emph{\textbf{V-CapsNet}} for the capsule network to improve its recognition rate.
    \item We propose a novel architectural block \emph{\textbf{NTT}} (\emph{Nontrainable-Trainable layer}) to improve the robustness of any CNN architecture against image degradation.
    \item We have explored the performance of different network architectures against adversarial attacks. In our experiments, we observe some unique properties of the proposed NTT layer against such perturbations.
    \item We have compiled two datasets -- (a) \emph{synthetic digits} and (b) \emph{natural images} for general benchmarking.
\end{itemize}

\section{Related Work}\label{sec:relatedwork}
Recent convolutional neural network architectures have achieved remarkable success rates in image classification. However, the recognition accuracy can be heavily affected by introducing various noise types. The primary reason is the difference in image quality between dataset samples and dynamically acquired application data. More specifically, while the networks are generally trained and validated on high-quality noise-free images, the captured image occasionally becomes noisy during inference on a real-time application, resulting in unexpected behavior even by a \emph{well-trained} network. Consequently, researchers have proposed different data preprocessing steps and network architectures to mitigate such issues with low-quality images. In \cite{ren2012coupled}, the authors use coupled kernel embedding to recognize faces from low-resolution images that also suffer from degradations due to high compression. In \cite{zou2012very}, the authors address the same problem of low-resolution face recognition by introducing discriminative constraints to achieve high-quality images for recognition. In \cite{basu2017learning}, a modified version of the MNIST dataset \cite{lecun1998gradient} is introduced by including synthetically generated noisy handwritten digits. The authors have used the dataset for representation learning with a novel framework involving probabilistic quadtrees and deep belief networks (DBN). A dataset of noisy faces is developed by \cite{karam2015quality} to benchmark the robustness of face recognition algorithms. However, the authors have not mentioned any network that achieves robust recognition results. Later, in \cite{tao2016multi}, the authors propose a joint kernel sparse representation-based classification algorithm that performs satisfactorily on the noisy face dataset \cite{karam2015quality}. Researchers \cite{ullman2016atoms} have also shown that the human vision system and convolutional neural networks process images differently. The authors have defined minimal recognizable configurations (MIRC) as the smallest cropped portion of the input image for which the class or action is perceivable to human vision. The subsequent experiments show that even state-of-the-art networks perform much worse and are not practically comparable to human vision. In \cite{dodge2016understanding, pei2018effects}, the authors have studied the effects of different degradation models on image classification by different deep architectures. However, the study is comparatively less exhaustive. It does not include noise models such as motion blur and salt-and-pepper or network architectures such as ResNet and CapsuleNet. Also, the authors \cite{dodge2016understanding, pei2018effects} have not proposed any \emph{defensive} mechanism in the network architectures to mitigate such performance degradations.

\section{Datasets and Experimental Setup}\label{sec:methodology}
For this study, we have compiled two datasets (synthetic digits and natural images) for training and evaluation. We have chosen six deep architectures (VGG-16 \cite{simonyan2014very}, VGG-19 \cite{simonyan2014very}, ResNet-50 \cite{he2016deep}, Inception-v3 \cite{szegedy2016rethinking}, MobileNet \cite{howard2017mobilenets} and CapsuleNet \cite{sabour2017dynamic}) with state-of-the-art recognition rates on the ImageNet dataset (ILSVRC) \cite{deng2009imagenet} to examine the robustness against six noise models (Gaussian white, Gaussian color, salt-and-pepper, Gaussian blur, motion blur and JPEG compression). To analyze a particular architecture against a specific noise model, we first train the network with the original noise-free training split. Then we evaluate the classification accuracy on the validation split multiple times. Each time, we degrade the validation images with gradually increasing perturbations following the noise model. This section briefly discusses the datasets, network architectures and degradation models used in this study.

\subsection{Datasets}
\textbf{Synthetic Digits:} The MNIST dataset \cite{lecun1998gradient} is used as a standard benchmark dataset for many network architectures in literature. However, due to the small image dimension of $28 \times 28$, the dataset is not ideal for benchmarking deep architectures with several convolution layers. For this reason, we introduce a more complex synthetic numerals dataset suitable for evaluating deep networks. We select 16 distinct fonts for 10 English numerals (0-9). Each numeral is rendered with a random color, rotation (-30$^\circ$ to +30$^\circ$), and size (30 to 240 pixels). Also, each numeral is superimposed on a random image taken from the COCO dataset \cite{lin2014microsoft} for additional complexity in the background. The dataset consists of 12000 images (10000 training samples + 2000 validation samples) equally distributed over 10 digits.

\textbf{Natural Images:} The ImageNet dataset \cite{deng2009imagenet} is a large-scale modern standard for evaluating deep neural networks. State-of-the-art CNNs are generally trained and validated on 1000 object classes from ImageNet. More recently, the capsule networks \cite{sabour2017dynamic} report very high recognition rates for image classification. However, due to the nature of the novel dynamic routing algorithm, training and validating capsule networks on 1000 classes are extremely difficult. Therefore, to study capsule networks alongside CNNs, a dataset with a smaller number of classes is more suitable. A potential candidate is the CIFAR-10 dataset \cite{krizhevsky2009learning}, which consists of 10 object classes. However, CIFAR-10 includes only low-resolution $32 \times 32$ images, making it a poor choice similar to MNIST for benchmarking very deep networks. For these reasons, we have compiled a dataset of high-resolution real-world images from 8 diverse classes. The dataset contains 6899 image samples of airplanes (727), cars (968), cats (885), dogs (702), flowers (843), fruits (1000), motorbikes (788) and persons (986). We have used 5724 images for training and 1175 images for validation in our experiments.

Fig. \ref{fig:dataset_samples} shows a few image samples from the proposed \emph{synthetic digits} and \emph{natural images} datasets.

\subsection{Deep neural network architectures}
\textbf{VGG-16:} The architecture \cite{simonyan2014very} consists of 16 layers having 138.4 million parameters. The network significantly improves large-scale image classification over all previous works, achieving the first discernable recognition performance on ILSVRC.

\textbf{VGG-19:} The architecture \cite{simonyan2014very} is a deeper and improved variant of VGG-16, consisting of 19 layers with 143.7 million parameters.

\textbf{ResNet-50:} Linear stacking of layers results in an exponentially large parameter space leading to difficulty in optimization. In \cite{he2016deep}, the authors have introduced \emph{residual learning} to address this issue. The architecture uses \emph{skip connections} to mitigate the vanishing gradient problem, leading to easier optimization during training. The proposed approach helps to construct very deep networks without exploding the parameter space. ResNet-50 consists of 50 layers having only 25.6 million parameters. The architecture is 2.6 times deeper with 82.8\% lesser parameters than VGG-19.

\textbf{Inception-v3:} The architecture \cite{szegedy2016rethinking} is motivated by the human vision system and introduces \emph{inception modules}, consisting of multiple depthwise concatenated convolution filters to extract multiscale image features. The network is 48 layers deep with 27.2 million parameters.

\textbf{MobileNet:} The architecture \cite{howard2017mobilenets} uses depthwise separable convolution layers to achieve a lightweight design with 28 layers having only 4.2 million parameters. The network design makes a trade-off between recognition accuracy and inference speed to achieve acceptable performance on embedded systems with limited computational resources.

\textbf{CapsuleNet:} In \cite{sabour2017dynamic}, the authors propose a novel \emph{dynamic routing} algorithm with \emph{capsule networks}. A \emph{capsule} is a group of neurons whose activity vector represents the instantiation of parameters corresponding to an object class. The architecture removes pooling layers in favor of dynamic routing between capsules to achieve translational invariance. The network consists of 6 layers with 8.2 million parameters.

The default configurations for the selected CNNs are specified for 1000 classes, and the same for CapsuleNet is 10 classes. Therefore, the baseline architectures are slightly adjusted for a direct and fair comparison. For the CNNs, all convolution layers in the \emph{backbone} are retained. However, the default fully-connected layers at the \emph{top} are replaced with three sequential fully-connected layers having 1024, 1024 and $n$ nodes, respectively. Here, $n$ specifies the number of object classes. So, $n = 10$ for \emph{synthetic digits} dataset and $n = 8$ for \emph{natural images} dataset. A dropout of 0.5 is applied between the first two fully-connected layers to minimize \emph{overfitting}. For the CapsuleNet, we introduce two additional convolution layers to reduce the input size before passing through the main architecture. Both layers use $9 \times 9$ convolution kernel and ReLU activation. The first convolution layer contains 256 filters and uses stride = 2. The second convolution layer contains 128 filters and uses stride = 1.

\subsection{Degradation models}

\begin{figure}[t]
    \includegraphics[width=\linewidth]{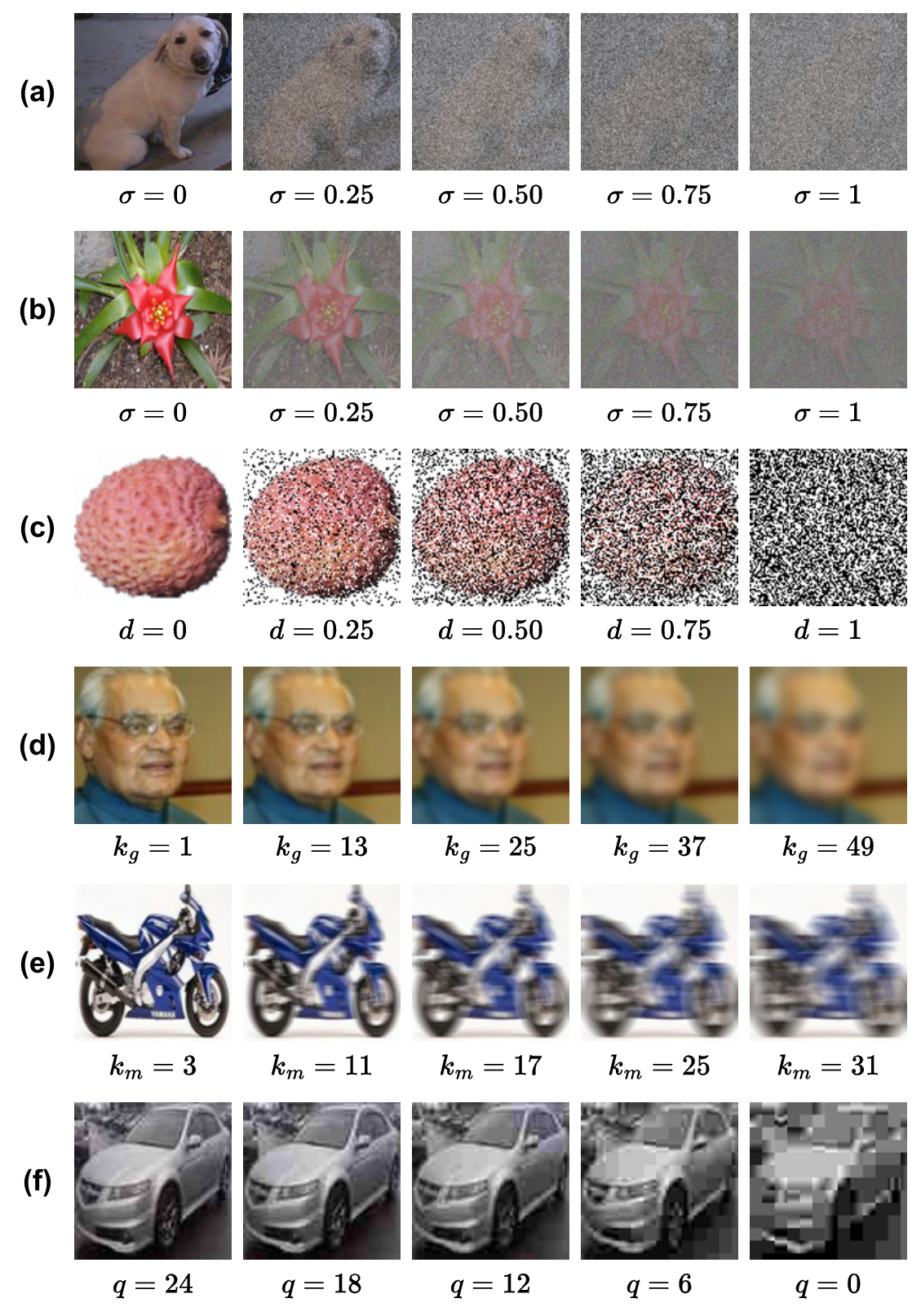}
    \caption{Examples of different image degradations with gradually increasing noise. (a) Gaussian white noise, (b) Gaussian color noise, (c) salt-and-pepper noise, (d) Gaussian blur, (e) motion blur, (f) JPEG compression (quality).}
    \label{fig:degradation_samples}
\end{figure}

\begin{figure*}[t]
    \includegraphics[width=\linewidth]{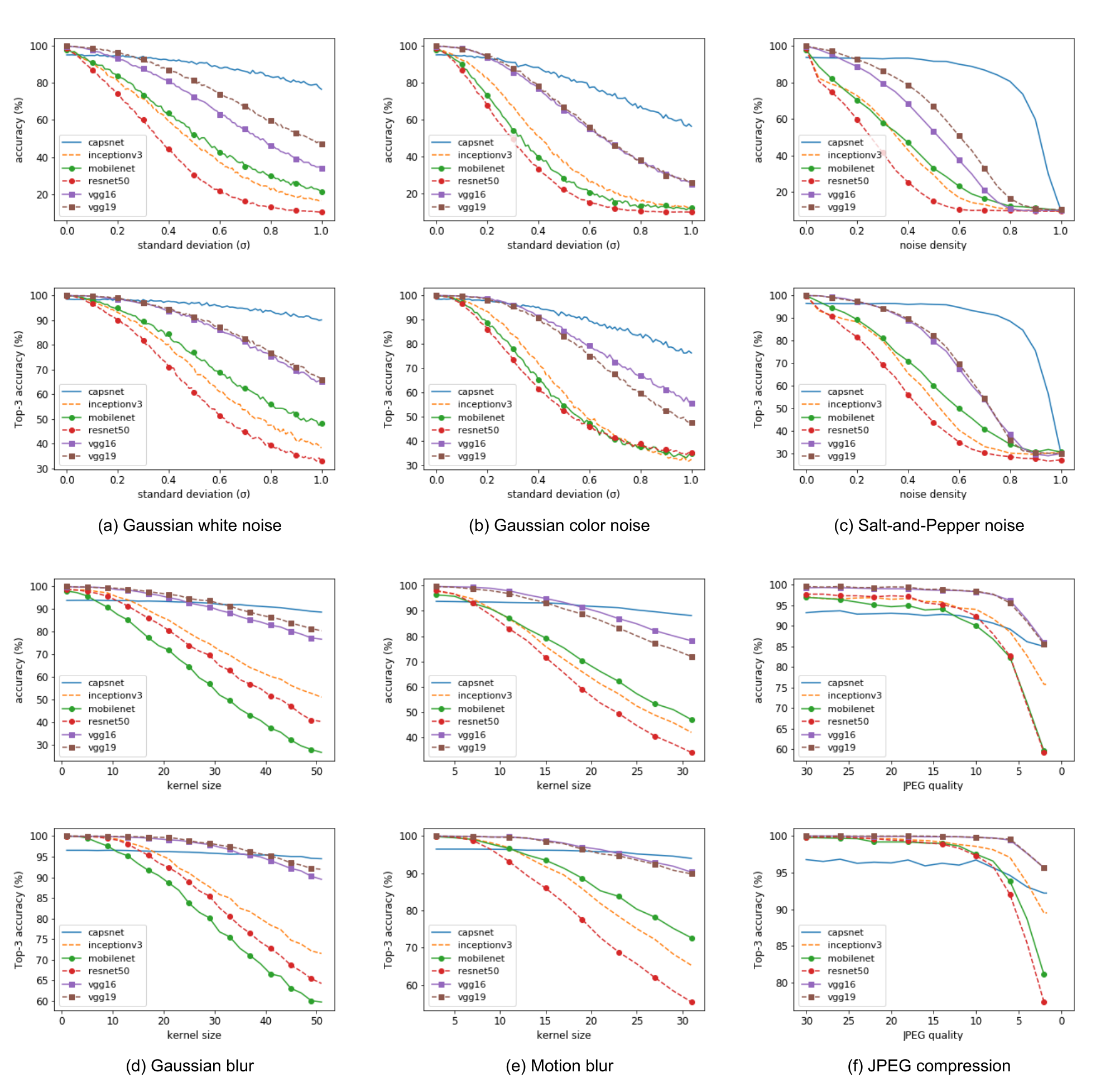}
    \caption{Comparison of classification accuracies of the CNN architectures against different image degradation models on the \textbf{synthetic digits} dataset.}
    \label{fig:results_sd}
\end{figure*}

\begin{figure*}[t]
    \includegraphics[width=\linewidth]{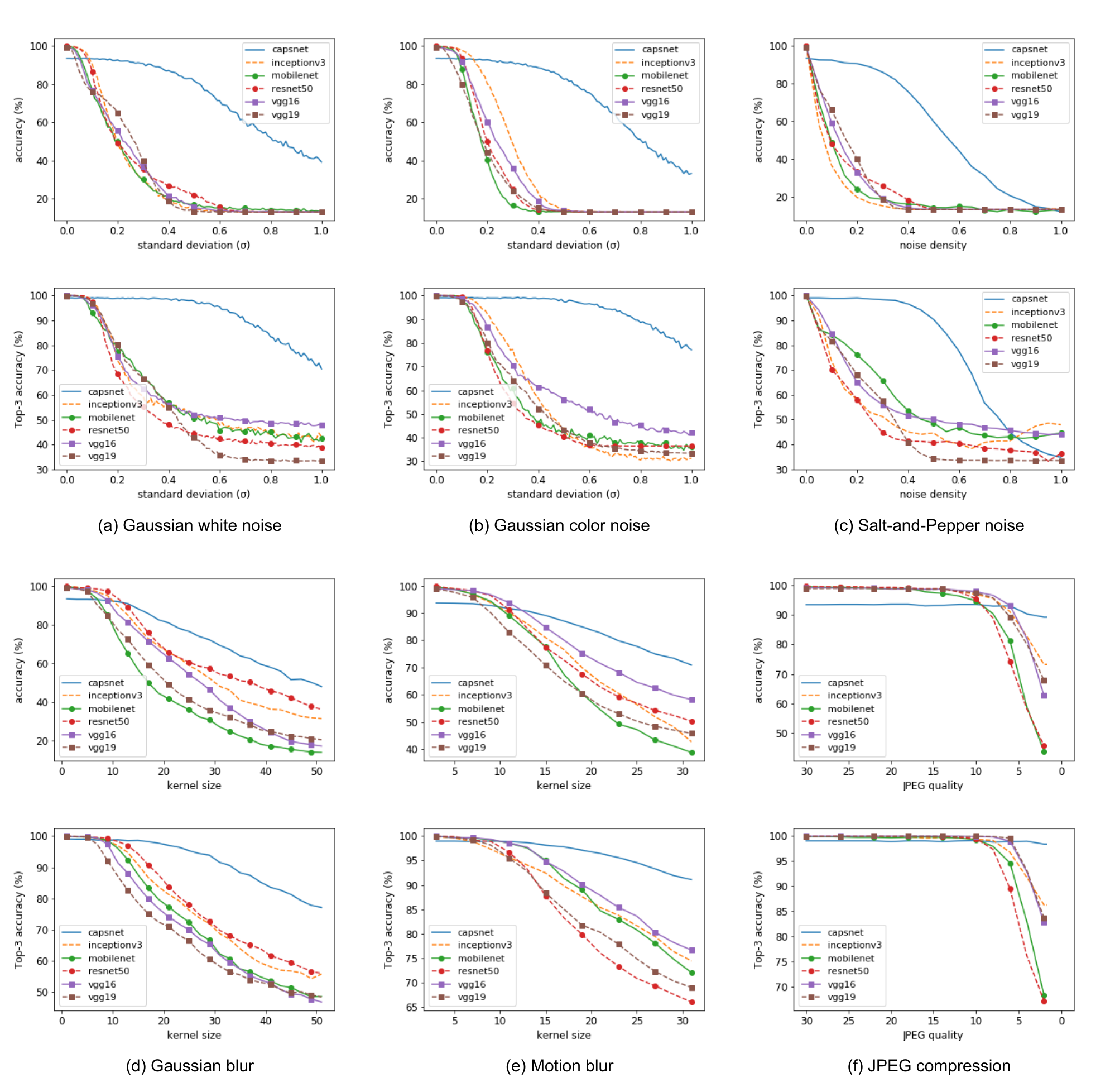}
    \caption{Comparison of classification accuracies of the CNN architectures against different image degradation models on the \textbf{natural images} dataset.}
    \label{fig:results_ni}
\end{figure*}

\begin{figure*}[t]
    \includegraphics[width=\linewidth]{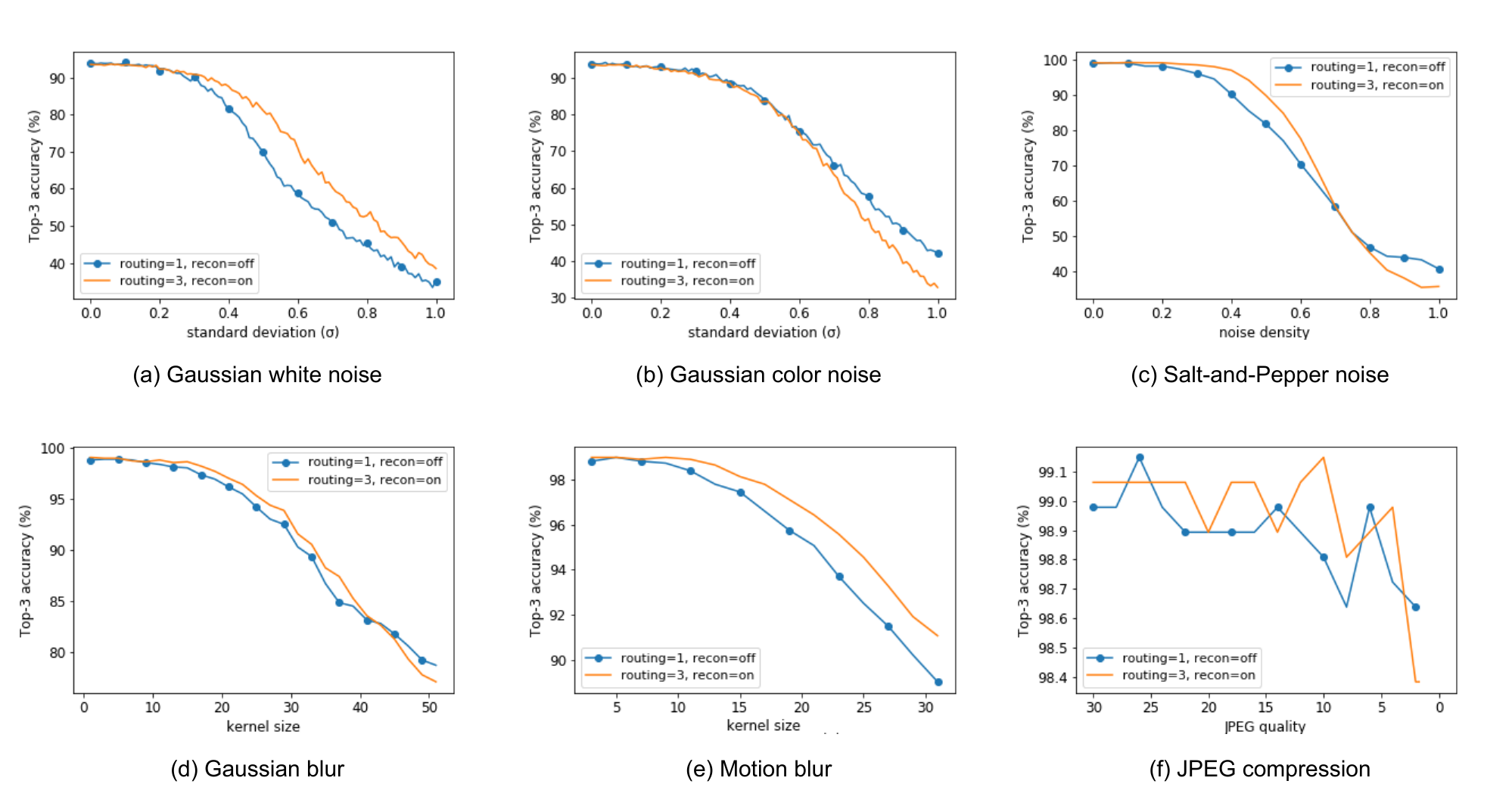}
    \caption{Comparison of classification accuracies of the CapsuleNet architectures having different hyperparameter configurations against different image degradation models.}
    \label{fig:results_capsulenet_routing_recon}
\end{figure*}

\textbf{Gaussian white noise:} Additive Gaussian noise arises when transmitting a digital image through a noisy channel or during image acquisition with a poor-quality sensor. In this study, we introduce synthetic Gaussian white noise into an image sample by adding a noise matrix of the same dimension as the image. The values of the noise matrix are sampled from a Gaussian distribution of zero mean and standard deviation $\sigma \ge 0$. We also keep the sampled values fixed across every channel of the noise matrix to generate synthetic Gaussian \emph{white} noise. In our experiments, we vary $\sigma$ from 0 to 1 with an interval of 0.1 for gradually increasing degradation of the image samples.

\textbf{Gaussian color noise:} We introduce synthetic Gaussian color noise similar to Gaussian white noise, but in this case, the sampled values are not kept fixed across the channels of the noise matrix. Here, we sample the values independently for each channel to generate synthetic Gaussian \emph{color} noise. In our experiments, we vary $\sigma$ from 0 to 1 with an interval of 0.1 for gradually increasing degradation of the image samples.

\textbf{Salt-and-pepper noise:} Salt-and-pepper is an impulse noise caused due to sparse but intense perturbations which randomly replace original pixel values in an image with white (\emph{salt}) or black (\emph{pepper}) pixels. We define the fraction of degraded pixels as the noise density $d$. For example, $d = 0.1$ indicates 10\% pixels of the image are degraded with salt-and-pepper noise. In our experiments, we vary $d$ from 0 to 1 with an interval of 0.1 for gradually increasing degradation of the image samples.

\textbf{Gaussian blur:} Gaussian blur roughly approximate blurring due to defocus or postprocessing. In this study, we introduce gradually increasing zero mean Gaussian blur by varying the kernel size $k_g$ from $3 \times3 $ to $51 \times 51$ with an interval of 2. The standard deviation of the blur is estimated as $\sigma_g = 0.3 * ((k_g - 1) * 0.5 - 1) + 0.8$.

\textbf{Motion blur:} During image acquisition, poor camera stabilization or fast object movements cause motion blur. In this study, we introduce gradually increasing horizontal motion blur by varying the kernel size $k_m$ from $1 \times 1$ to $31 \times 31$ with an interval of 2.

\textbf{JPEG compression:} Image compression is essential for processing, storing and transmitting digital images. JPEG is the most commonly used lossy image compression technique. Therefore, we have closely studied JPEG compression as a degradation model in this work. In our experiments, we introduce gradually increasing degradation by varying the output quality parameter $q$ of the JPEG encoder from 30 to 0 with an interval of 2. A lower quality value indicates better compression but higher visual perturbations in the images.

In Fig. \ref{fig:degradation_samples}, we show examples of different image degradation models with gradually increasing noise.

\section{Results and analysis}\label{sec:results}
In this study, we gradually increase the image perturbation using a selected noise model and measure the validation accuracy of each network on the degraded images. We also measure the top-3 accuracy for further investigating the performance of the networks against image degradation. Fig. \ref{fig:results_sd} and Fig. \ref{fig:results_ni} show the results of these experiments on the \emph{synthetic digits} and \emph{natural images} datasets, respectively.

\subsection{Effects of image degradations on CNN architectures}
The experimental results show decreasing classification accuracy for every CNN architecture with the increasing variance of the additive Gaussian noise. For both Gaussian white and color noise models, VGG architectures exhibit the lowest performance decay among the conventional CNNs. Interestingly, CapsuleNet shows the highest robustness against additive Gaussian noise.

Although the recognition accuracy of CapsuleNet is slightly lower than other architectures in the absence of any impulse noise, CapsuleNet begins to outperform other CNNs when noise density $d \ge 0.2$. CapsuleNet retains its superior recognition rate up to $d \le 0.8$. After this point, the classification accuracy rapidly falls to similar values as generic CNNs.

Both Gaussian blur and motion blur degrade the performance of CNNs. However, VGG networks generally perform better than deeper architectures such as ResNet, Inception and MobileNet. Interestingly, the recognition rate of CapsuleNet is not significantly impacted by blurring, and for heavily blurred images CapsuleNet effectively outperforms other CNNs.

JPEG compression does not exhibit any significant degradation of classification accuracy for the networks when the quality parameter $q > 20$, as the visible structure of the images remains nearly unaltered. Below this threshold, $0 \le q \le 20$, the recognition rate of ResNet and MobileNet degrades sharply. However, VGG, Inception and CapsuleNet architectures perform significantly better in this range.

To analyze the potential effects of the \emph{dynamic routing} algorithm \cite{sabour2017dynamic} against image degradations, we test CapsuleNet with two different configurations -- (a) with single routing iteration and optimizing only marginal loss; (b) with 3 routing iterations and optimizing reconstruction loss combined with marginal loss. Fig. \ref{fig:results_capsulenet_routing_recon} shows a slight improvement in recognition rate with higher routing iterations and the reconstruction loss as a regularizer.

\subsection{Analysis of network depth in CapsuleNet architecture}
\begin{figure}[t]
    \includegraphics[width=\linewidth]{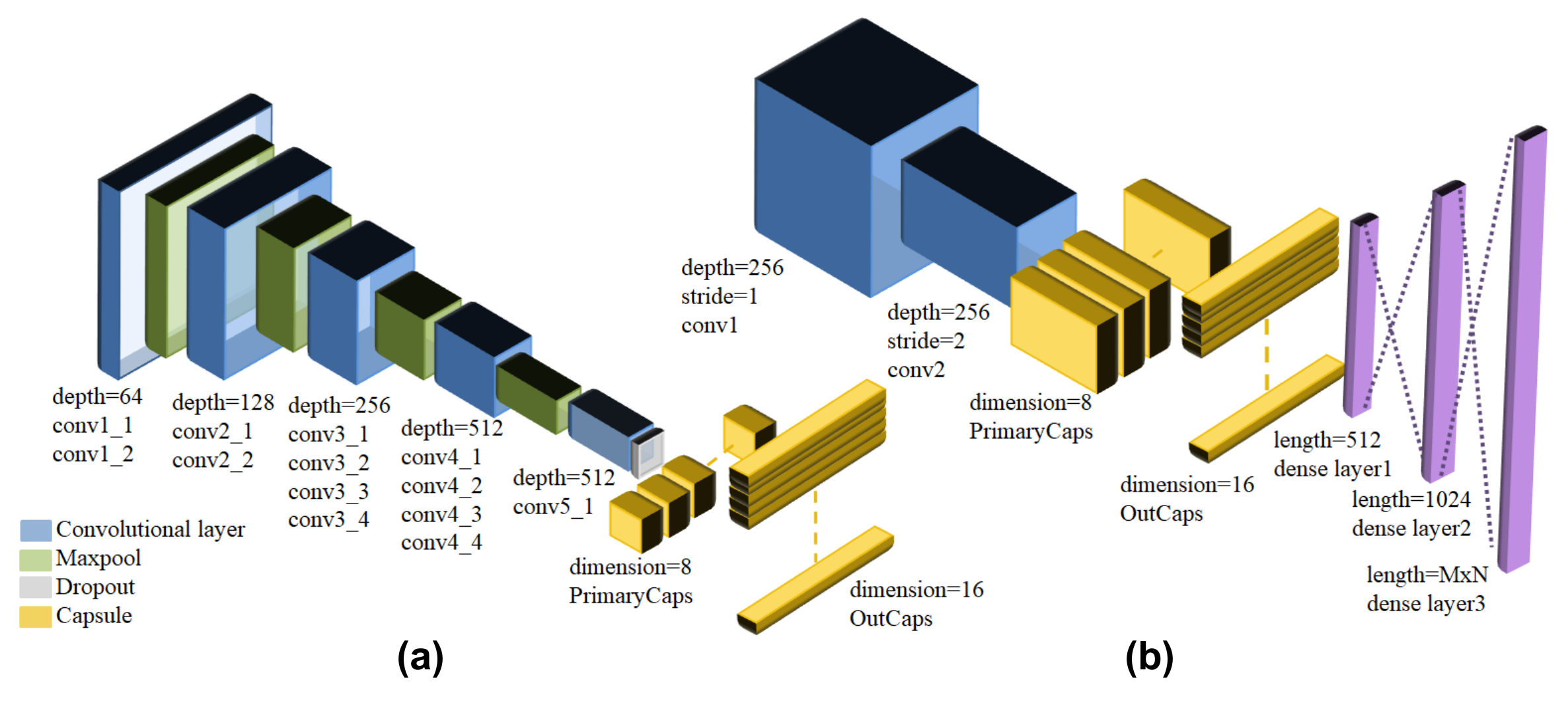}
    \caption{Two different capsule network architectures. (a) proposed V-CapsNet, (b) baseline CapsuleNet.}
    \label{fig:vcapsnet}
\end{figure}

\begin{figure*}[t]
    \includegraphics[width=\linewidth]{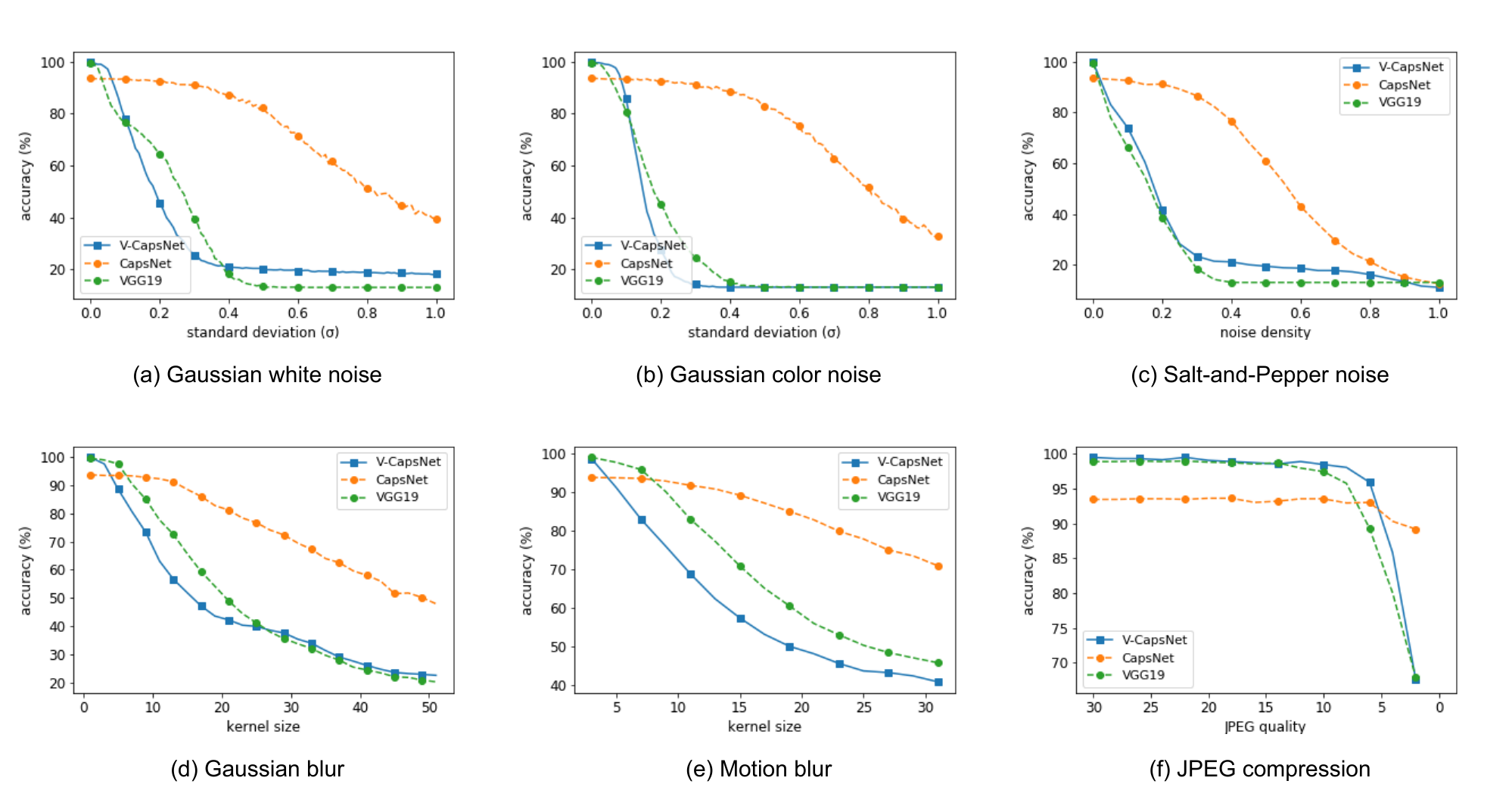}
    \caption{Comparison of classification accuracies of VGG-19, CapsuleNet and V-CapsNet against different image degradation models.}
    \label{fig:results_vcapsnet}
\end{figure*}

In our study, CapsuleNet shows significantly higher robustness against image degradation than conventional deep CNNs. However, state-of-the-art deep CNNs achieve better recognition accuracy than CapsuleNet for noise-free samples of all datasets. To improve the baseline performance of CapsuleNet, we introduce a novel fusion architecture \emph{\textbf{V-CapsNet}} by integrating convolution layers of the VGG-19 network into CapsuleNet. We replace the initial convolution layers in CapsuleNet architecture with the convolution layers (up to the first layer of the fifth block) from VGG-19. The resulting feature maps are passed into the \emph{PrimaryCaps} layer of CapsuleNet, which uses 256 convolution kernels of size $3 \times 3$ with stride = 2. The PrimaryCaps layer has $32 \times 7 \times 7$ capsule outputs where each capsule is an 8-dimensional vector. The final layer has one 16-dimensional capsule per class, identical to the original architecture \cite{sabour2017dynamic}. We drop the decoder of the baseline CapsuleNet to reduce the number of trainable parameters and optimize the network by minimizing the marginal loss only. In our experiments, the proposed V-CapsNet fusion architecture achieves 99.83\% validation accuracy on the \emph{natural images} dataset, improving the baseline performance of CapsuleNet by 6.2\%. Fig. \ref{fig:vcapsnet} shows the architecture of the proposed V-CapsNet alongside the baseline CapsuleNet. Interestingly, V-CapsNet is more sensitive to image degradation than CapsuleNet and VGG-19. In Fig. \ref{fig:results_vcapsnet}, we show the classification accuracies of VGG-19, CapsuleNet and V-CapsNet against different image degradation models. Although V-CapsNet initially has a much better recognition accuracy, the performance sharply falls below the baseline CapsuleNet with increasing noise.

\begin{figure*}[t]
    \includegraphics[width=\linewidth]{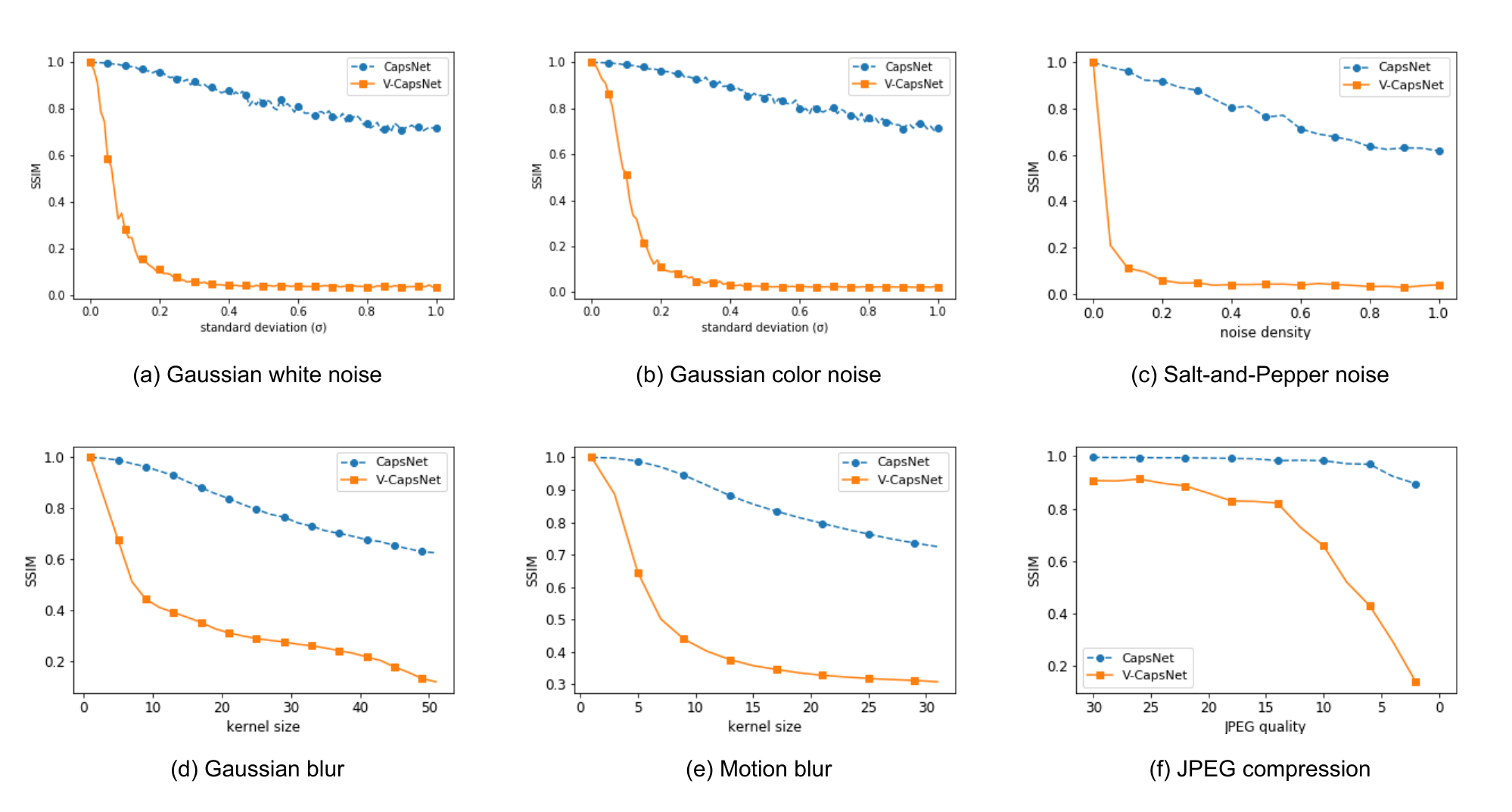}
    \caption{Comparison of SSIM between the final output feature maps with and without introducing noise for baseline CapsuleNet and V-CapsNet.}
    \label{fig:results_vcapsnet_ssim}
\end{figure*}

While investigating the reasons behind the resilience of baseline CapsuleNet over V-CapsNet against image degradations, we observe that both capsule layers and shallowness of the network depth contribute to the noise robustness in the baseline architecture. We validate our hypothesis by independently measuring the change in output feature maps from the final convolution layer of each network. The structural similarity index (SSIM) \cite{wang2004image} is used as the evaluation metric. We measure the SSIM of the feature maps with and without introducing noise in the input image samples. The change in SSIM is found to be much slower for baseline CapsuleNet than V-CapsNet, as shown in Fig. \ref{fig:results_vcapsnet_ssim}. Therefore, we believe the noise robustness is related to the depth of the network. Apparently, Deeper networks are more sensitive to image degradations. For conventional CNNs, deeper architectures have shown significant performance gain over shallower architectures. However, CapsuleNet uses a novel \emph{dynamic routing} algorithm to achieve classification accuracy close to deep CNNs while using a much shallower architecture. V-CapsNet improves the baseline performance of CapsuleNet by increasing the network depth and thereby exposing the architecture more toward image degradations.

\section{Improving robustness of deep architectures against image degradations}\label{sec:nttlayer}
Although deep network architectures are more susceptible to noise, they facilitate better classification accuracy by learning complex nonlinearity in the feature space. CapsuleNet achieves decent recognition rates with a more noise-tolerant shallow architecture. Attempting to improve the performance of CapsuleNet with a deeper fusion architecture leads to higher classification accuracy but lower robustness against image degradations. So, a relevant research question arises whether it is possible to construct deeper architectures with higher recognition rates without trading off robustness toward image degradations. One possible way to address the issue is by including noisy samples in the training dataset. However, this approach has several limitations. Firstly, including noisy samples for every probable noise model results in an extremely large sample space. Also, estimating every possible noise type with varying levels is nearly impossible. For example, generating synthetic degraded image samples with linear blur kernels is straightforward, but producing such samples for all probable nonlinear blur kernels is difficult. Therefore, it is desirable to construct networks that are inherently robust against image degradations.

\subsection{Proposed method for improving robustness against noise}

\begin{figure}[t]
    \includegraphics[width=\linewidth]{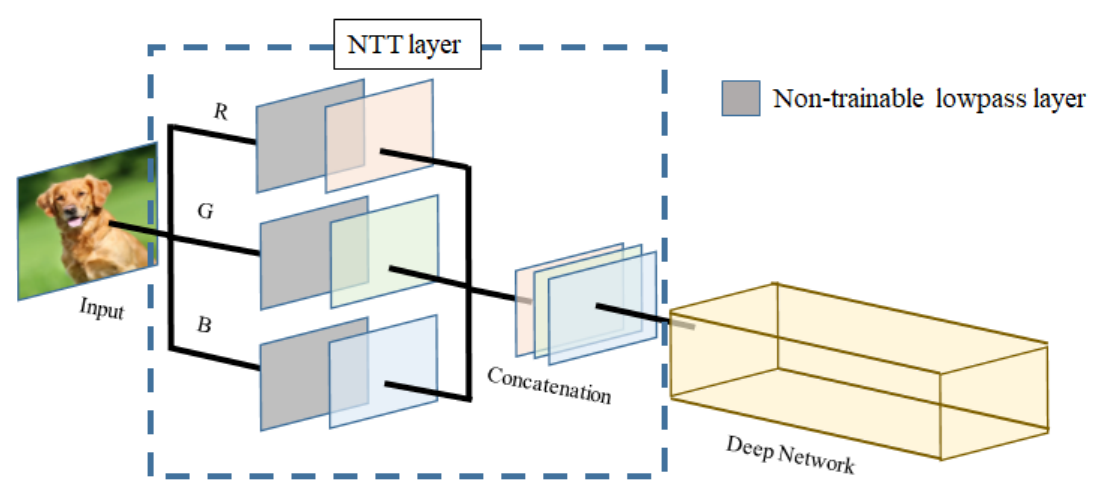}
    \caption{The proposed \emph{\textbf{nontrainable-trainable (NTT)}} layer for improving the robustness of a network architecture against image degradations.}
    \label{fig:nttlayer}
\end{figure}

\begin{figure*}[t]
    \includegraphics[width=\linewidth]{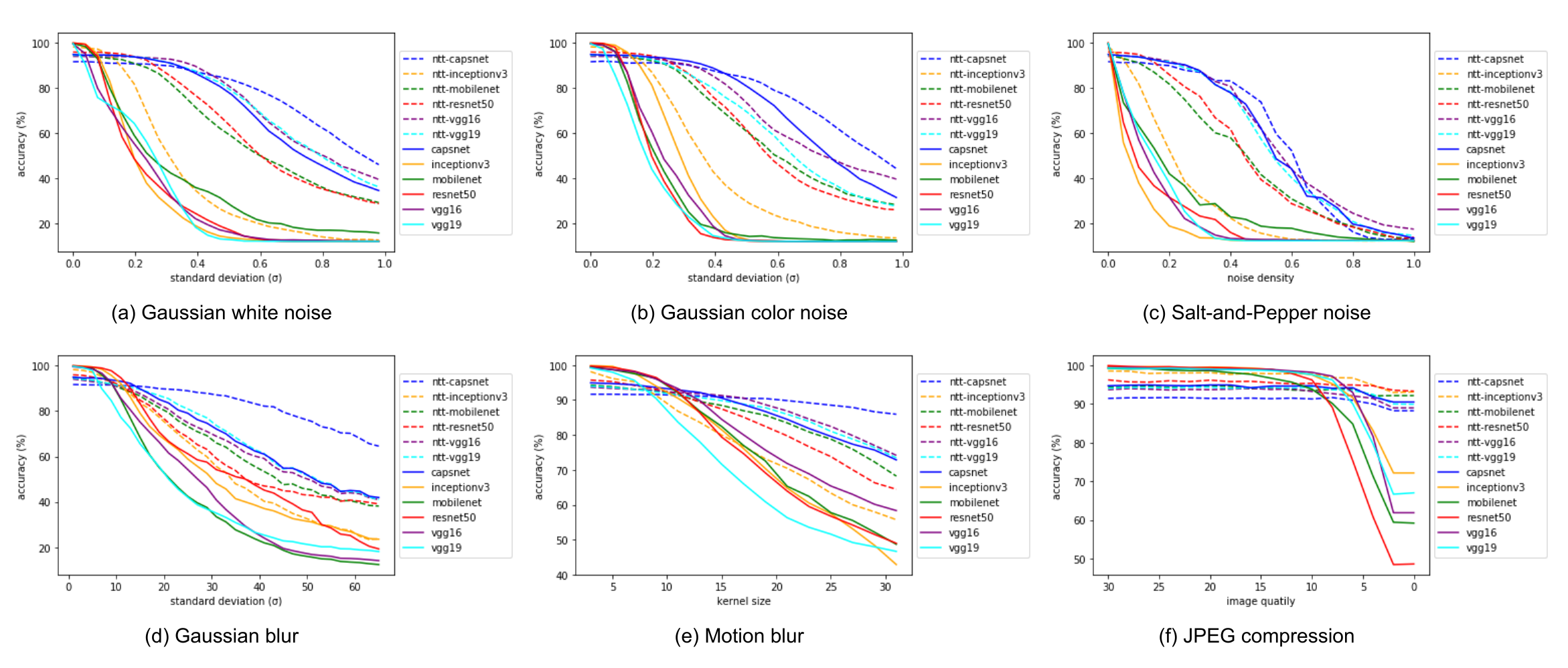}
    \caption{Comparison of classification accuracies of the CNN architectures against different image degradation models with and without using the \textbf{NTT} layer.}
    \label{fig:results_nttlayer}
\end{figure*}

\begin{figure*}[t]
    \includegraphics[width=\linewidth]{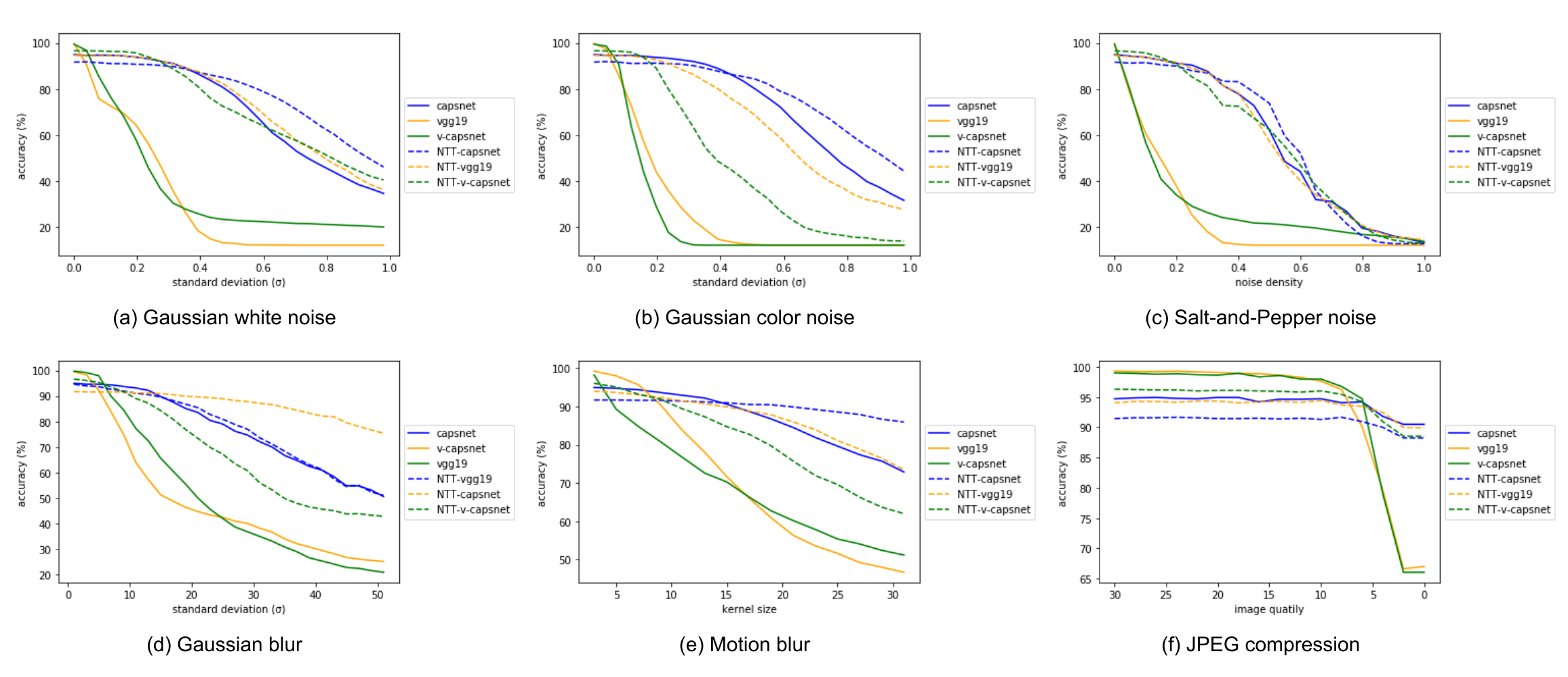}
    \caption{Comparison of classification accuracies of VGG-19, CapsuleNet and V-CapsNet against different image degradation models with and without using the \textbf{NTT} layer.}
    \label{fig:results_nttlayer_vcapsnet}
\end{figure*}

\begin{figure}[t]
    \includegraphics[width=\linewidth]{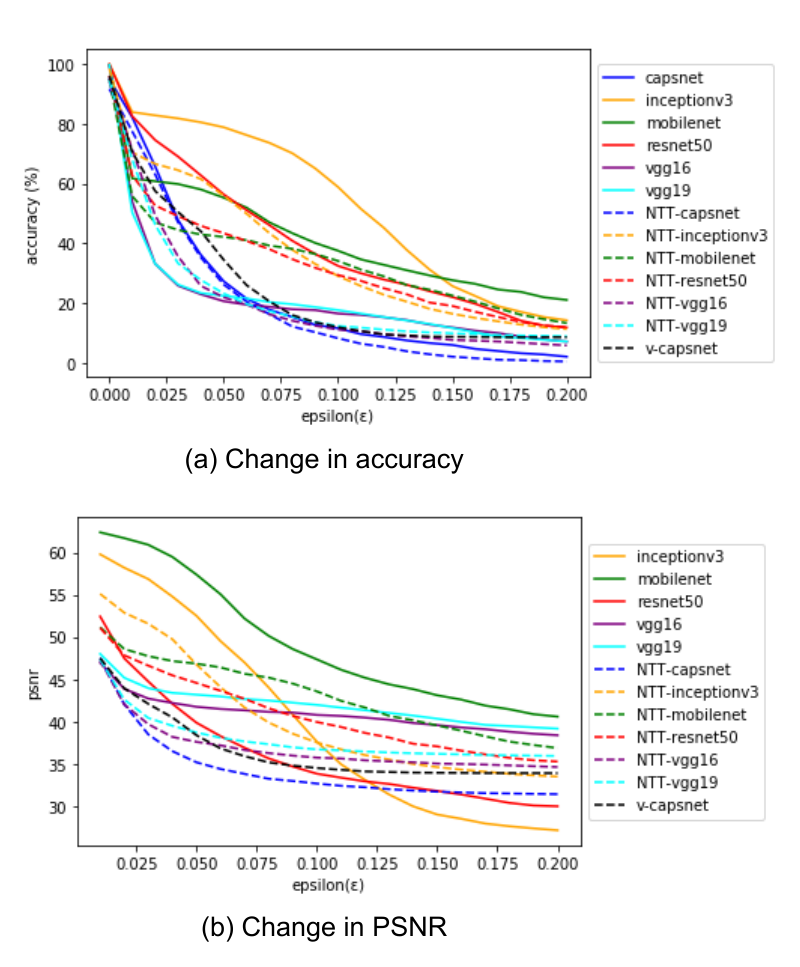}
    \caption{Comparison of the CNN architectures against adversarial attack.}
    \label{fig:results_nttlayer_fgsm}
\end{figure}

We introduce a novel architectural block with depthwise filtering to improve the robustness of a network against image degradations without incurring any significant performance bottleneck. The proposed \emph{\textbf{nontrainable-trainable (NTT)}} layer consists of one depthwise nontrainable filter followed by one depthwise trainable filter. The nontrainable layer uses the weights of a lowpass filter. Fig. \ref{fig:nttlayer} shows the structure of a composite NTT layer. Both the nontrainable and trainable layers use one single-channel filter with unit stride. The motivation behind the NTT layer is to learn the invariance toward subtle changes introduced by noising and denoising. When we train a network with standard good-quality image samples but attempt to perform inference on denoised images, the recognition accuracy is heavily affected as several subtle structural details get modified during denoising. The NTT layer mitigates this issue by helping the network learn the invariance toward such subtle structural artifacts.

Fig. \ref{fig:results_nttlayer} shows the comparison of classification accuracies of the CNN architectures against different image degradation models with and without using the NTT layer. We observe a significant improvement in robustness for every network against image degradations when injecting the NTT layer into architecture. A point to note is that this increased resilience against noise using NTT layers also causes a slight decrease in the baseline accuracy of the CNNs on noise-free images. However, the bottleneck in baseline accuracy is significantly smaller than the general performance gain by the CNNs. For example, introducing the NTT layer in VGG-19 trades off 6\% baseline accuracy in favor of up to 68\% improvement in the presence of Gaussian color and salt-and-pepper noise. The drop in maximum baseline accuracy and the amount of performance gain vary depending on the network architecture, but the general behavior remains the same. The trade-off between accuracy and robustness is controllable by adjusting network depth and filter size in the NTT layer.

We also analyze the efficacy of the NTT layer for the proposed V-CapsNet fusion architecture. Fig. \ref{fig:results_nttlayer_vcapsnet} shows the comparison of classification accuracies of VGG-19, CapsuleNet and V-CapsNet against different image degradation models with and without using the NTT layer. In most cases, the NTT layer pushes the maximum recognition accuracy of V-CapsNet close to VGG-19 while improving the noise robustness close to the baseline CapsuleNet.

\subsection{Defense against adversarial perturbations}
Perceptible degradation of input image quality is one of the prime reasons for the poor performance of CNNs. However, researchers have shown that carefully estimated imperceptible changes can drastically impact the classification accuracy of CNNs. These \emph{adversarial perturbations} can be introduced into the input image samples in several ways \cite{rauber2017foolbox}. Even a trivially crafted adversarial attack \cite{kurakin2016adversarial} can significantly reduce the performance of a network. To study the robustness of CNNs against adversarial attacks, we introduce gradually increasing perturbations into the image samples using the fast gradient sign method (FGSM) \cite{kurakin2016adversarial}. FGSM estimates the gradient of the loss with respect to the intensity at each pixel location, followed by modifying the pixel intensity by an amount $\epsilon$ such that the loss increases. In our study, we use FGSM for \emph{untargeted} adversarial attack, which updates the pixel intensities such that the network misclassifies the input image to any other class. The value of $\epsilon$ affects the perceptible change in the modified image samples. We compare the robustness of the networks using $\epsilon$ as the evaluation metric because a higher value of $\epsilon$ is required to force a more robust network toward incorrect inference. In our experiments, CapsuleNet and VGG architectures perform poorly, while ResNet-50, Inception-v3, MobileNet and the proposed V-CapsNet perform well.

A study on the adversarial robustness of different architectures with NTT layers reveals an intriguing observation. As shown in Fig. \ref{fig:results_nttlayer_fgsm}(a), injecting NTT layers causes a faster decrease in recognition accuracy, making it easier to create adversarial perturbations for such networks. However, Fig. \ref{fig:results_nttlayer_fgsm}(b) shows that the average peak signal-to-noise ratio (PSNR) of the perturbed images rapidly drops for a network with an NTT layer, indicating a higher visual degradation in image quality. Therefore, even imperceptible adversarial perturbations with low $\epsilon$ are more likely to cause visible distortions during the attack, leading to easier visual identification of an adversarially perturbed image sample. An important point to note is that the performance of a network against adversarial attacks does not correlate to its robustness against image degradation models.

\section{Conclusion}\label{sec:conclusion}
This paper investigates the effects of different image degradation models on CNN architectures for image classification. While every network follows a general trend of decreasing recognition rate with gradually increasing noise, we observe that deeper networks are more sensitive toward image perturbations. The baseline CapsuleNet is remarkably robust against all image degradation models and shows excellent resilience for salt-and-pepper noise and blurring. To further improve the performance of CapsuleNet, we introduce a fusion architecture \emph{V-CapsNet} by replacing the feature extraction layers in the baseline CapsuleNet with that of VGG-19. While the fusion architecture achieves the highest classification accuracy among all the networks in our experiments, the performance also decays faster than other networks. We hypothesize that robustness against noise depends on the depth of the architecture, and simply going deeper into the architecture improves the recognition rate by trading off stability against perturbations. To address this dilemma, we propose a novel architectural block \emph{NTT}, which improves the robustness of any convolutional network against image degradations. We also find it difficult to comment on the robustness of CNNs against adversarial attacks by drawing a direct analogy from their performance on perceptible image distortions. It is worthwhile to explore future scopes of constructing noise-robust deep networks that can achieve high recognition accuracy.

\section*{Acknowledgements}
We would like to thank \href{https://www.nvidia.com}{NVIDIA\textsuperscript{\textregistered} Corporation} for providing a TITAN X GPU through the \href{https://academicgrants.nvidia.com/academicgrantprogram/s/welcome}{GPU Grant Program}.

\bibliographystyle{IEEEtran}
\bibliography{references}





\end{document}